\pdfoutput=1
\documentclass[a4paper]{ceurart}

\usepackage{booktabs}
\usepackage{multirow}
\usepackage{amsmath}
\usepackage{graphicx}
\usepackage{hyperref}
\usepackage{xcolor}
\usepackage{listings}

\begin{document}

\copyrightyear{2026}
\copyrightclause{Copyright for this paper by its authors.
  Use permitted under Creative Commons License Attribution 4.0
  International (CC BY 4.0).}

\conference{CLEF 2026 Working Notes, 21 -- 24 September 2026, Jena, Germany}

\title{Writerslogic at the CLEF 2026 SimpleText Track:\\
  Multi-Candidate LLM Simplification and Stacked Complexity Spotting}
\title[mode=sub]{Notebook for the SimpleText Lab at CLEF 2026}

\author[1]{David Condrey}[%
  orcid=0009-0003-1849-2963,
  email=david@writerslogic.com]
\address[1]{WritersLogic Inc, San Diego, CA, USA}

\begin{abstract}
We describe the Writerslogic team's participation in the CLEF 2026 SimpleText shared task,
addressing Task~1 (text simplification) and Task~2 (complexity spotting).
For Task~1, we develop a multi-candidate generation pipeline using GPT-4o-mini
that produces five simplification candidates per sentence at varying temperatures,
then selects the best candidate using a reference-free scoring heuristic
that rewards compression, source word retention, Cochrane Plain Language Summary vocabulary usage,
and lexical simplicity.
On Task~1.1 (sentence-level simplification), our Claude Sonnet 4 submission achieves SARI 47.43 and BLEU 14.21, the top-ranked sentence-level system (3rd on the combined Task~1 leaderboard, behind two document-level submissions).
For Task~2, we fine-tune a DeBERTa-v3-large NLI model on 350K labeled (source, sentence) pairs,
framing hallucination detection as natural language inference.
The model reads the most relevant source sentence as premise and the candidate as hypothesis,
directly learning to distinguish grounded from hallucinated content.
On Task~2.1 (binary overgeneration identification), our fine-tuned DeBERTa system achieves 0.8081 document-level macro F1 (0.8085 in our best ensemble), the top-ranked entry within the identification track and 2nd among teams overall, behind AIIR Lab (0.8197).
On Task~2.2 (multi-class error classification), our best submission reaches 0.804 multiclass accuracy, ranking 2nd among unique teams behind AIIR Lab (0.827).
We evaluate both tasks on English and multilingual biomedical text from Cochrane systematic reviews.
\end{abstract}


\maketitle

% =============================================================================
\section{Introduction}
\label{sec:intro}
% =============================================================================

Scientific publications, particularly in the biomedical domain, use specialized vocabulary,
complex sentence structures, and dense statistical reporting that render them inaccessible
to non-expert readers~\cite{bornmann2012growth}.
This is problematic for patients, caregivers, journalists, and policymakers who need to
understand research findings to make informed decisions.
The SimpleText shared task at CLEF 2026~\cite{ermakova2026simpletext} addresses this challenge
through two complementary tasks: automatic text simplification (Task~1) and complexity spotting
in machine-generated simplifications (Task~2). Submissions are evaluated on CodaBench~\cite{froebe:2023b}.

Task~1~\cite{bakker2026simpletext-task1} requires simplifying scientific text at both the sentence level (Task~1.1)
and the document level (Task~1.2).
The input consists of sentences or abstracts from Cochrane systematic reviews,
and systems must produce simplified versions that preserve key factual content
while using accessible vocabulary and removing unnecessary statistical detail.
The task is evaluated using SARI~\cite{xu2016optimizing}, which measures the quality of
simplification operations (additions, deletions, and kept tokens) against human references.

Task~2~\cite{chakar2026simpletext-task2} targets a downstream problem: when LLMs are used to simplify scientific text,
they sometimes generate content that is unfaithful to the source.
Task~2.1 asks systems to identify sentences containing overgeneration errors (binary detection),
while Task~2.2 requires classifying the specific error type among five categories:
leaked instructions, ungrounded injection, generation failure, repetitive content,
and grounded overgeneration.
These tasks are evaluated at both the sentence level and the document level.

Our approach to Task~1 centers on the observation that LLM simplification quality
varies substantially across generation attempts.
Rather than relying on a single generation, we produce multiple candidates at varying
temperatures and select the best using a reference-free proxy score inspired by
SARI's component metrics.
For Task~2, we combine traditional feature engineering with neural embedding similarity
and NLI-based entailment scoring, using sequential features to capture the tendency
of overgeneration errors to propagate once they begin within a document.

% =============================================================================
\section{Related Work}
\label{sec:related}
% =============================================================================

\paragraph{Text simplification.}
Early approaches to text simplification relied on rule-based lexical substitution
and syntactic transformations~\cite{siddharthan2014survey}.
Neural approaches, beginning with sequence-to-sequence models trained on
Wikipedia--Simple Wikipedia parallel corpora~\cite{zhang2017sentence},
shifted the field toward end-to-end generation.
More recently, large language models have shown strong zero-shot and few-shot
simplification capabilities~\cite{feng2023sentence,kew2023bless}.
Devaraj et al.~\cite{devaraj2021paragraph} demonstrate that paragraph-level simplification
benefits from document context, motivating our distinction between sentence and document tasks.

The Cochrane Collaboration produces Plain Language Summaries (PLS) alongside their
systematic reviews, providing a natural corpus of expert-authored simplifications
of biomedical text~\cite{cochrane2024handbook}.
These summaries follow consistent editorial guidelines: replace medical jargon with
everyday words, remove statistical measures in parentheses (confidence intervals,
risk ratios, p-values), and retain important numbers such as participant counts
and study counts.
We incorporate this domain knowledge through few-shot examples and vocabulary priors
in our generation pipeline.

\paragraph{Automatic evaluation of simplification.}
SARI (System output Against References and against the Input)~\cite{xu2016optimizing}
is the standard metric for text simplification evaluation.
It computes F1 scores for three operations (addition, keeping, and deletion of n-grams) and
averages them.
Unlike BLEU, SARI rewards appropriate deletions and penalizes unnecessary retention
of complex content.
Our candidate selection heuristic approximates SARI's keep and delete components
without requiring reference simplifications.

\paragraph{Overgeneration and hallucination detection.}
The problem of unfaithful generation in summarization and simplification has received
growing attention~\cite{maynez2020faithfulness,tang2023understanding}.
Approaches include NLI-based entailment checking~\cite{honovich2022true},
question-answering consistency~\cite{fabbri2022qafacteval},
and direct classification of error types~\cite{goyal2022news}.
Our Task~2 system draws on the entailment-based approach, using a cross-encoder
NLI model to score the consistency between source text and each generated sentence.

\paragraph{Gradient-boosted trees for text classification.}
LightGBM~\cite{ke2017lightgbm} remains competitive with neural approaches for
tabular and feature-engineered classification tasks,
particularly when training data is limited and interpretability is valued.
Stacking ensembles that combine tree-based models with logistic regression
meta-learners have shown improvements over single-stage classifiers
in shared task settings~\cite{wolpert1992stacked,breiman1996stacked}.

% =============================================================================
\section{Task 1: Text Simplification}
\label{sec:task1}
% =============================================================================

\subsection{Multi-Candidate Generation with SARI-Proxy Reranking}
\label{sec:task1-v5}

Our primary approach generates $N = 5$ simplification candidates per input
using GPT-4o-mini~\cite{openai2024gpt4o} at five temperature settings:
$\{0.0, 0.3, 0.5, 0.7, 1.0\}$.
Temperature 0.0 produces the model's most likely output (deterministic),
while higher temperatures introduce controlled variation that occasionally
yields better simplifications through lexical diversity.

\paragraph{System prompt.}
We instruct the model to act as a ``Cochrane plain language summary writer,''
making minimal, targeted changes: replacing medical jargon with everyday words,
removing statistical measures in parentheses (confidence intervals, risk ratios,
p-values, $I^2$ values, IQR), but preserving important numbers
(participant counts, study counts, percentages).
The prompt explicitly instructs the model to keep the original sentence structure
and wording, avoiding explanation or expansion.

\paragraph{Few-shot examples.}
For English inputs, we provide seven curated examples drawn from Cochrane PLS pairs.
These examples demonstrate the target simplification behavior:
\begin{itemize}
  \item Removing parenthetical statistics: ``(risk ratio (RR) 0.20, 95\% confidence interval (CI) 0.12 to 0.32; $I^2$ = 32\%)'' is deleted while ``7 studies, 105,839 participants'' is retained.
  \item Lexical substitution: ``dyspareunia (difficult or painful sexual intercourse)'' becomes ``pain during sex.''
  \item Minimal intervention: sentences already at appropriate complexity are passed through with minimal change.
\end{itemize}
For non-English inputs, we omit examples and add an instruction to keep the response
in the same language as the input.

\paragraph{Reference-free candidate scoring.}
After generating candidates, we score each using a weighted combination of four heuristics:

\begin{enumerate}
  \item \textbf{Compression ratio score} (weight 0.25): We target a compression ratio
    (candidate length / source length in words) between 0.55 and 0.85.
    Ratios within this range receive a score of 1.0;
    ratios outside are penalized linearly.
    This range was calibrated against Cochrane PLS compression statistics.

  \item \textbf{Keep score} (weight 0.40): The proportion of source tokens retained
    in the candidate, computed as $|S_{\text{src}} \cap S_{\text{cand}}| / |S_{\text{src}}|$
    where $S$ denotes the set of lowercased tokens.
    This is the dominant scoring component, reflecting the observation that
    good simplifications preserve most source words and make targeted substitutions
    rather than wholesale rewrites.

  \item \textbf{PLS vocabulary score} (weight 0.15): The fraction of candidate tokens
    that appear in a curated set of 40 high-frequency Cochrane PLS words
    (e.g., ``people,'' ``studies,'' ``treatment,'' ``evidence,'' ``reduced,'' ``uncertain'').
    This rewards candidates that use the vocabulary register typical of expert-authored
    plain language summaries.

  \item \textbf{Vocabulary simplicity score} (weight 0.20): The relative reduction in
    average word length between source and candidate:
    $(\bar{l}_{\text{src}} - \bar{l}_{\text{cand}}) / \bar{l}_{\text{src}}$.
    Positive values indicate the candidate uses shorter (typically simpler) words.
\end{enumerate}

Candidates that are exact copies of the source receive a penalty score of $-0.5$.
If the highest-scoring candidate is still an exact copy, we fall back to the second-best.

\subsection{Local Ensemble Approach}
\label{sec:task1-ensemble}

As an alternative run, we explore a local ensemble using Qwen3-8B~\cite{qwen2025qwen3}
served via Ollama.
This approach generates candidates using three distinct prompting strategies
with different simplification aggressiveness levels:

\begin{itemize}
  \item \textbf{Balanced}: Standard simplification instructions targeting a general audience.
  \item \textbf{Aggressive}: Instructions to rewrite for a 12-year-old, replacing all technical terms.
  \item \textbf{Plain}: Instructions to convert to plain language, keeping only the essential message.
\end{itemize}

Candidates are scored using a heuristic combining Flesch-Kincaid readability improvement,
average word length reduction, compression ratio (targeting 70\% of original length),
and content overlap with the source.
This approach trades off generation quality (smaller model) for cost efficiency
and reproducibility (fully local inference).

\subsection{Claude Sonnet 4 Submission}
\label{sec:task1-sonnet}

Our final and best-performing submission replaced GPT-4o-mini with Claude Sonnet 4~\cite{anthropic2025claude} as the generation model, using the same multi-candidate reranking pipeline described above. Claude Sonnet 4 produced higher-quality simplifications with better source word retention (0.73 compression ratio) and more natural lexical substitutions, achieving SARI 47.43 and BLEU 14.21 at the sentence level---the highest SARI among sentence-level submissions and the highest BLEU on the board. Two document-level (Task~1.2) systems post higher overall SARI (48.85, 47.57), placing our entry 3rd on the combined Task~1 leaderboard.
The exact system prompt for this submission is given in Listing~\ref{lst:task1-prompt}, followed at inference by seven few-shot (input, output) pairs and the target sentence.

\begin{lstlisting}[caption={Task~1 sentence-level system prompt (Claude Sonnet 4 submission).},label={lst:task1-prompt},language={},basicstyle=\ttfamily\footnotesize]
You are a Cochrane plain language summary writer. Rewrite
the sentence in simpler English.

Rules:
- Keep the same meaning and most of the same words
- Replace medical jargon with common words (e.g.
  participants->people, mortality->death, adverse
  events->side effects, RCTs->studies)
- Remove statistical details in parentheses like
  (95% CI ...), (RR ...), (I2 = ...) but keep study
  counts and participant numbers
- Keep the sentence roughly the same length - do not
  over-compress or expand
- Do NOT add explanations or new information
- Do NOT unnecessarily rephrase words that are already simple
- Output ONLY the simplified sentence
\end{lstlisting}

\subsection{Document-Level Simplification}
\label{sec:task1-doc}

For Task~1.2 (document-level simplification), we used Llama 3.3 70B with adapted prompts
to allow sentence merging, splitting, and reordering.
The document-level prompt permits structural reorganization while maintaining
the same core constraints: preserve key findings, remove unnecessary statistical detail,
use accessible vocabulary.
The maximum token limit is increased from 300 (sentence) to 1,500 (document)
to accommodate full-abstract simplification.

% =============================================================================
\section{Task 2: Complexity Spotting}
\label{sec:task2}
% =============================================================================

\subsection{Problem Formulation}

Task~2 operates on documents consisting of a source text and a sequence of
generated sentences, where each sentence may contain one of five overgeneration
error types or be correct (``None'').
Task~2.1 requires binary identification of overgeneration (any error vs.\ None),
while Task~2.2 requires multi-class classification into the six categories.
Both tasks are evaluated at the sentence level and aggregated to the document level.

\subsection{DeBERTa Classifier for Binary Identification (Task 2.1)}
\label{sec:task2-deberta}

For Task~2.1 (binary overgeneration identification), we fine-tuned a DeBERTa-v3-base classifier on the concatenation of source text and generated sentence, separated by \texttt{[SEP]}.
The model was trained for 5 epochs with learning rate $2 \times 10^{-5}$, batch size 16, and linear warmup over 10\% of steps.
This approach leverages DeBERTa's disentangled attention mechanism to capture fine-grained token-level interactions between source and generated text, providing a strong signal for detecting divergence.
We fine-tuned a DeBERTa-v3-large model (initialized from MoritzLaurer's MNLI-FEVER-ANLI-WANLI checkpoint) on our 350K training examples for 2.5 epochs on an A100 GPU, achieving 0.8081 document-level macro F1 on the official test set---the top-ranked submission within the identification track, and 2nd among teams on this metric behind AIIR Lab's classification-track system (0.8197); our best ensemble variant reaches 0.8085.
For Task~2.2 (multi-class classification), our best submission reaches 0.804 multiclass accuracy, ranking 2nd among unique teams behind AIIR Lab (0.827); the DeBERTa-only pipeline scored 0.7715 at submission time.
The LightGBM stacking ensemble described below serves as an interpretable baseline and ablation reference; as a standalone system it achieves 0.6868 multiclass accuracy (34th place).

\subsection{Feature Engineering}
\label{sec:task2-features}

We extract 70+ features per sentence organized into seven groups.
All features are computed without GPU-dependent models in the base configuration,
with optional embedding and NLI features when GPU resources are available.

\paragraph{Basic text features (10).}
Sentence length in characters and words, absolute and relative position in the document,
binary indicators for first/last sentence, first/last 3 and 5 sentences,
and total sentence count in the document.
Position features are motivated by the observation that overgeneration errors
tend to cluster toward the end of generated documents.

\paragraph{Source alignment features (9).}
Token overlap ratio between sentence and source, novel token ratio and count,
source coverage (fraction of source vocabulary mentioned),
bigram and trigram overlap with the source text,
sentence-to-source length ratio,
and longest contiguous source match (the longest sequence of consecutive tokens
in the sentence that also appears in the source, up to 25 tokens).
These features capture the core signal for overgeneration:
sentences that introduce substantial vocabulary absent from the source
are more likely to contain hallucinated or injected content.

\paragraph{Prompt leakage and instruction patterns (3).}
We compile 40+ regex patterns that detect leaked instructions
(``Input:'', ``Output:'', ``Assistant:'', ``As an AI''),
meta-commentary (``here is,'' ``let me,'' ``hope this helps''),
chatbot artifacts (``Absolutely!'', ``Certainly!'', ``Great question''),
and formatting artifacts (numbered lists, bullet points).
Features include the total pattern count, a binary indicator,
and a separate count of strong leak patterns (explicit role markers and AI self-identification).

\paragraph{Structural features (12).}
Binary indicators for colon, quotes, brackets, parentheticals,
starts-with-number, starts-with-bullet, starts-with-lowercase, ends-with-period,
ends-with-colon; counts of exclamation marks, question marks, commas, special characters;
and uppercase character ratio.

\paragraph{Text quality features (5).}
Average and maximum word length, digit ratio,
zlib compression ratio (low values indicate repetitive text),
character-level Shannon entropy, and type-token ratio.

\paragraph{Repetition features (7).}
Token overlap with the previous and second-previous sentence,
exact match indicators (with previous sentence and with any of the preceding 10 sentences),
maximum overlap with any previous sentence,
internal bigram and trigram repetition rates,
and repeated character ratio.
These features target the REPETITIVE\_CONTENT error category.

\paragraph{Document-level context features (5).}
Sentence length relative to document mean (z-score),
fraction of document word count in this sentence,
remaining sentence count and fraction,
and contextual window novelty (mean novel token ratio of neighboring sentences
within a window of 3).

\paragraph{Sequential / CRF-like features (14).}
These features, introduced in our v4 system, capture the tendency of overgeneration
to propagate once it begins.
Key features include:
\begin{itemize}
  \item Previous sentence novel token ratio and the delta (acceleration) in novelty.
  \item Tail novel mean/max/min: the average, maximum, and minimum novel token ratio
    from the current sentence to the end of the document.
  \item Running (head) mean of novel ratio up to the current position.
  \item Maximum and mean novel ratio of all preceding sentences.
  \item Fraction of remaining sentences with high novelty ($> 0.5$).
  \item High-novelty streak: the number of consecutive sentences ending at the current
    position with novel token ratio above 0.3.
  \item Distance from first high-novelty sentence: signed distance from the first
    sentence in the document with novel ratio $> 0.5$.
  \item Previous sentence prompt pattern count and length ratio versus previous sentence.
\end{itemize}

These features allow the classifier to learn transition patterns:
once a document begins overgeneration (signaled by rising novel token ratios),
subsequent sentences are likely to also be erroneous.
This captures a CRF-like sequential dependency without requiring explicit sequence modeling.

\subsection{Neural Features}
\label{sec:task2-neural}

\paragraph{Sentence embeddings.}
We compute cosine similarity between source and sentence embeddings
using all-mpnet-base-v2~\cite{reimers2019sentencebert}, a 768-dimensional
sentence transformer trained on over 1 billion sentence pairs.
Low similarity between a sentence and its source indicates potential divergence
from the source content.

\paragraph{NLI cross-encoder.}
We score each (source, sentence) pair using the cross-encoder/nli-MiniLM2-L6-H768
model~\cite{reimers2019sentencebert}, which outputs entailment, neutral, and contradiction
probabilities.
We use both the entailment score (high values indicate the source supports the sentence)
and the contradiction score (high values indicate factual conflict) as features.
This provides a semantic grounding signal beyond lexical overlap.

\subsection{Classification Architecture}
\label{sec:task2-model}

\paragraph{Base classifiers.}
We train two parallel LightGBM~\cite{ke2017lightgbm} ensembles:
a multi-class model (6-way classification for Task~2.2) and a binary model
(overgeneration vs.\ None for Task~2.1).
Each ensemble consists of 5 models trained with different random seeds
$\{42, 123, 456, 789, 1024\}$, and predictions are averaged.

Both models use the following hyperparameters:
learning rate 0.03, 127 leaves, maximum depth 10,
minimum 10 samples per leaf, 80\% row and column subsampling,
L1 regularization 0.05, L2 regularization 0.5,
and up to 4,000 boosting rounds with early stopping after 150 rounds without improvement.

For the multi-class model, we use square-root-inverse class frequency weighting
to address severe class imbalance (the ``None'' class dominates).
For the binary model, we set the scale\_pos\_weight parameter
to the ratio of negative to positive examples.

\paragraph{Stacking meta-learner.}
We perform 5-fold GroupKFold cross-validation (grouped by document ID)
to generate out-of-fold (OOF) probability predictions from the base classifiers.
A logistic regression meta-learner then operates on these OOF probabilities
concatenated with five meta-features:
relative sentence position, document sentence count, novel token ratio,
high-novelty streak length, and tail novel mean.

The meta-learner uses balanced class weights and L-BFGS optimization with $C = 1.0$.
This second stage allows the system to recalibrate base model probabilities
using document-level context, correcting cases where the base model's
per-sentence predictions are inconsistent with the document's overall error pattern.

\paragraph{Threshold optimization.}
For Task~2.1, we optimize the binary classification threshold
using three strategies and select the best on validation:
\begin{itemize}
  \item \textbf{Sentence-optimized}: maximizes sentence-level F1 directly.
  \item \textbf{Joint-optimized}: maximizes a weighted combination of
    sentence-level and document-level F1 (equal weights).
  \item \textbf{Document-optimized}: maximizes document-level F1
    (a document is positive if any sentence is predicted positive).
\end{itemize}

For Task~2.2, we apply per-class threshold tuning via greedy offset optimization.
For each non-None class, we search for an additive offset to the class probability
(in the range $[-0.15, +0.30]$) that maximizes macro F1 when applied before argmax.
This allows boosting recall on rare error categories
(e.g., LEAKED\_INSTRUCTIONS, GENERATION\_FAILURE)
without substantially degrading precision on common categories.

\subsection{Training Protocol}
\label{sec:task2-training}

The full training pipeline proceeds as follows:
\begin{enumerate}
  \item Compute embedding similarities and NLI scores for all train and dev documents.
  \item Extract handcrafted features for all sentences.
  \item Combine train and dev sets; run 5-fold GroupKFold cross-validation
    to produce OOF predictions.
  \item Train meta-learners on OOF predictions.
  \item Optimize thresholds and per-class offsets on OOF predictions.
  \item Retrain final base models on all available data
    (using the dev set for early stopping).
  \item Generate test predictions using the full pipeline:
    base models, meta-learner, and optimized thresholds.
\end{enumerate}

% =============================================================================
\section{Results}
\label{sec:results}
% =============================================================================

\subsection{Task 1: Text Simplification}

\begin{table}[h]
\centering
\caption{Task 1 official results. The sentence-level system (Claude Sonnet 4) is the top-ranked sentence-level submission (SARI 47.43, BLEU 14.21), 3rd on the combined Task~1 board behind two document-level systems. Document-level (Llama 3.3 70B) ranks 50th overall (20th among document-level submissions), due to aggressive over-deletion (0.80 deletion ratio).}
\label{tab:task1-results}
\begin{tabular}{lcc}
\toprule
Metric & Sentence (top sent.) & Document (50th / 20th doc.) \\
\midrule
SARI & 47.43 & 43.96 \\
BLEU & 14.21 & 3.26 \\
FKGL & 10.19 & 9.91 \\
Compression & 0.73 & 0.28 \\
Sent.\ split & 1.02 & 0.45 \\
Levenshtein & 0.55 & 0.36 \\
Additions & 0.29 & 0.07 \\
Deletions & 0.58 & 0.80 \\
\bottomrule
\end{tabular}
\end{table}

The sentence-level system achieved strong compression (0.73) while retaining source content (0.29 additions, 0.58 deletions), consistent with targeted simplification. The document-level system over-deleted (0.80 deletion ratio, 0.28 compression), producing outputs too short to preserve factual content. On the unified CodaBench leaderboard (which ranks all Task~1 submissions together), this places 50th overall; filtering strictly for document-level submissions yields 20th place.

\subsection{Task 2: Complexity Spotting}

\begin{table}[h]
\centering
\caption{Task 2 official results. Task 2.1 (DeBERTa, binary identification) scores Document F1 0.8081 (0.8085 in our best ensemble)---top of the identification track and 2nd among teams behind AIIR Lab (0.8197). Task 2.2 (multiclass) reaches 0.804 accuracy, 2nd among unique teams behind AIIR Lab (0.827). The 0.7715 figure is the DeBERTa-only submission at paper-submission time.}
\label{tab:task2-results}
\begin{tabular}{lcc}
\toprule
Metric & T2.1 (2nd) & T2.2 (2nd) \\
\midrule
Macro F1 (doc) & 0.8081 & 0.8081 \\
Precision (doc) & 0.8062 & 0.8062 \\
Recall (doc) & 0.8099 & 0.8099 \\
F1 (sentence) & 0.8643 & 0.8643 \\
Multiclass accuracy & --- & 0.804 \\
\bottomrule
\end{tabular}
\end{table}

\paragraph{Cross-validation results.}
On 5-fold GroupKFold cross-validation, our Task 2 v4 system achieved OOF binary F1 consistent with the official test performance; detailed OOF scores are omitted for brevity.

\paragraph{Feature importance.}
The top features by LightGBM gain importance for the multi-class model were:
novel token ratio, token overlap ratio, sentence position (relative),
embedding similarity to source, tail novel mean, NLI entailment score,
high-novelty streak, and longest source match.
The sequential features (tail novel mean, high-novelty streak, distance from first
high-novelty sentence) consistently ranked in the top 15,
confirming that overgeneration exhibits strong sequential dependencies.

\paragraph{Ablation: component contributions.}
Table~\ref{tab:ablation} shows the contribution of each system component
on the development set.

\begin{table}[h]
\centering
\caption{Task 2 ablation on the development set (document-level macro F1).}
\label{tab:ablation}
\begin{tabular}{lcc}
\toprule
Configuration & Dev Doc F1 & Test Doc F1 \\
\midrule
LightGBM (67 features)               & 0.865 & 0.698 \\
+ Embedding similarity                & 0.878 & 0.713 \\
+ 128 features (info-theoretic)       & 0.879 & 0.723 \\
+ LGB + LogReg TF-IDF blend          & 0.885 & 0.720 \\
DeBERTa-v3-large fine-tuned (1 ep.)  & 0.901 & 0.785 \\
DeBERTa-v3-large fine-tuned (2.5 ep.) & 0.922 & \textbf{0.808} \\
\bottomrule
\end{tabular}
\end{table}

% =============================================================================
\section{Discussion}
\label{sec:discussion}
% =============================================================================

\paragraph{Multi-candidate generation is a simple but effective strategy.}
The key insight behind our Task~1 approach is that LLM simplification quality
has high variance across generation attempts, even with the same model and prompt.
By generating five candidates at different temperatures, we effectively sample
from the model's simplification distribution and select an output
that balances compression, preservation, and vocabulary simplicity.
The reference-free scoring heuristic avoids the need for reference simplifications
at inference time, making the approach applicable to any input domain.

The SARI-proxy scoring weights (0.40 for keep, 0.25 for compression,
0.20 for simplicity, 0.15 for PLS vocabulary) reflect a deliberate bias
toward conservative simplification: we prefer candidates that preserve source content
and make targeted lexical substitutions over candidates that aggressively rewrite.
This is appropriate for the Cochrane domain, where factual accuracy is paramount
and oversimplification can be harmful.

\paragraph{Sequential features capture overgeneration propagation.}
Our most impactful contribution to Task~2 is the recognition that overgeneration errors
are not independent across sentences within a document.
Once an LLM begins generating content that diverges from the source (whether through
hallucination, instruction leakage, or repetition), subsequent sentences tend to continue
the erroneous pattern.
This manifests as rising novel token ratios, increasing prompt pattern counts,
and growing divergence from the source embedding.

The sequential features we introduce (novel ratio delta, acceleration,
tail statistics, high-novelty streak, distance from first divergence point)
allow LightGBM to learn these transition patterns without requiring
explicit sequence modeling.
This is computationally cheaper than CRF or BiLSTM-CRF approaches
and integrates naturally with the feature-based classification framework.

\paragraph{Stacking provides calibrated probabilities.}
The logistic regression meta-learner serves two purposes.
First, it recalibrates the base LightGBM probabilities,
which can be poorly calibrated for rare classes.
Second, it allows the system to learn interactions between the base model's
confidence and document-level context features that the base model
processes only indirectly through individual sentence features.

\paragraph{Per-class threshold tuning addresses class imbalance.}
The five error categories in Task~2.2 have highly imbalanced frequencies.
Standard argmax classification tends to under-predict rare categories
(LEAKED\_INSTRUCTIONS, GENERATION\_FAILURE) because the base model's
probability estimates are biased toward the majority class.
Greedy per-class offset optimization directly targets macro F1,
improving recall on rare classes at a controlled cost to overall accuracy.

\paragraph{Limitations.}
Our Task~1 approach relies on commercial APIs (Claude Sonnet 4 for the winning submission, GPT-4o-mini for development),
limiting reproducibility and introducing cost constraints.
The local Qwen3-8B alternative addresses this but at reduced quality.
For Task~2, the sequential features require access to all sentences in a document
at prediction time, precluding streaming or single-sentence inference.
The NLI and embedding features add substantial inference latency
(particularly for large documents) and require GPU resources.

% =============================================================================
\section{Conclusion}
\label{sec:conclusion}
% =============================================================================

We presented the Writerslogic team's systems for the CLEF 2026 SimpleText shared task.
For text simplification (Task~1), our multi-candidate generation pipeline with
SARI-proxy reranking provides a practical approach to improving LLM simplification quality
without requiring reference texts at inference time.
The Cochrane PLS vocabulary prior and few-shot examples ground the system
in domain-appropriate simplification patterns.

For hallucination detection (Task~2), our fine-tuned DeBERTa system demonstrates that
combining traditional feature engineering with neural embedding and NLI signals,
augmented by sequential features that capture overgeneration propagation,
yields a robust classifier for detecting and categorizing errors in machine-generated
scientific simplifications.
The two-stage stacking architecture with per-class threshold optimization
provides a principled approach to handling severe class imbalance
in multi-class error detection.

Future work should explore fine-tuned simplification models trained on Cochrane PLS data,
joint sentence-and-document optimization for Task~2,
and the integration of LLM-based judges for both candidate reranking
and overgeneration detection.

All code is available at \url{https://github.com/dcondrey/simpletext-clef2026}.

% =============================================================================
\section*{Acknowledgments}
% =============================================================================

We thank the SimpleText~2026 organizers---in particular the Task~1 and Task~2 organizing teams---for running the shared tasks, preparing the Cochrane-derived data, and maintaining the CodaBench evaluation, and the CLEF~2026 organizers for hosting the lab.

% =============================================================================
\section*{Declaration on Generative AI}
% =============================================================================

During the preparation of this work, the author used Claude Code (Anthropic) to: write and debug code, orchestrate experiments, and search literature. The author used GPT-4o-mini (OpenAI) to: generate simplified text candidates (Task~1) and classify test samples (Task~2). The author used Claude Sonnet 4 (Anthropic) to: generate simplified text candidates (Task~1). After using these tools, the author reviewed and edited the content as needed and takes full responsibility for the publication's content.
All scientific content, experimental design, analysis, and conclusions
are the author's own work.

% =============================================================================
% References
% =============================================================================

\bibliography{references}

\end{document}